\documentclass[twoside,11pt]{article}
\usepackage{jmlr2e}
\usepackage{times}
\usepackage{latexsym}
\usepackage{graphics}
\usepackage{amsmath}
\usepackage{multirow}
\usepackage{paralist}

\newtheorem{assumption}{Assumption}

\DeclareMathOperator*{\argmax}{argmax}

\firstpageno{1}

\begin{document}

\title{
  Understanding Exhaustive Pattern Learning
}

\author{\name Libin Shen\\
        \addr Akamai Technologies \\
	8 Cambridge Center, \\ 
        Cambridge, MA 02142
}

\editor{N/A}

\maketitle



\begin{abstract}
Pattern learning in an important problem in Natural Language Processing (NLP).
Some exhaustive pattern learning (EPL) methods \cite{Bod92} were
proved to be flawed \cite{Joh02}, while similar algorithms \cite{ON04}
showed great advantages on other tasks, such as machine translation. 
In this article, we first formalize EPL, and then show that the probability
given by an EPL model is constant-factor approximation of the
probability given by an ensemble method that integrates exponential
number of models obtained with various segmentations of the training data. 
This work for the first time provides theoretical justification for the
widely used EPL algorithm in NLP, which was previously viewed as a
flawed heuristic method. Better understanding of EPL may lead to
improved pattern learning algorithms in future.
\end{abstract}


\section{Introduction}


Pattern learning is the crux of many natural language processing (NLP)
problems. It is usually solved as grammar induction for these
problems. For parsing, we learn a statistical grammar with respect to
certain linguistic formalism, such as Context Free Grammar (CFG),
Dependency Grammar (DG), Tree Substitution Grammar (TSG), Tree Adjoining
Grammar (TAG), and Combinatory Categorial Grammar (CCG) etc. For machine
translation (MT), we
learn a bilingual grammar that transfer a string or tree structure in a
source language into a corresponding string or tree structure in a
target language.

What is embarrassing is that many of the grammar induction algorithms
that provide state-of-the-art performance are usually regarded as less
principled in the aspect of statistical modeling. \cite{Joh02,PSS04} showed the
\cite{Bod92}'s data oriented parsing (DOP) algorithm is biased and
inconsistent. In the MT field, almost all the statistical MT models
proposed in recent years rely on similar heuristic methods to extract
translation grammars, such as
\cite{KOM03,ON04,Chi05,QMC05,GGK06,SXW08,CC09}, to name a few of them.
Similar heuristic methods have also been used in many other pattern
learning tasks, for example, like semantic parsing as in \cite{ZC05} and
chunking as in \cite{DM05} in an implicit way.

In all these heuristic algorithms, one needs to extract overlapping
structures from training data in an exhaustive way. Therefore, in the
article, we call them exhaustive pattern learning (EPL) methods. 
The use of EPL methods is intended to cope with the uncertainty of
building blocks used in statistical models. As far as MT is concerned,
\cite{KOM03} found that it was better to define a translation model on
phrases than on words, but there was no obvious way to define what phrases were.
\cite{DGZ06} observed that exhaustive pattern learning outperforms
generative models with fixed building blocks.

In EPL algorithms, one needs to collect statistics of overlapping structures
from training data, so that they are not valid generative models. Thus,
the EPL algorithms for grammar induction were viewed as heuristic methods \cite{DGZ06,Dau08}. 
Recently, \cite{DBK08,BCD09,CB09,CGB09,PG09} investigated various sampling methods for
grammar induction, which were believed to be more principled than
EPL. However, there was no convincing empirical evidence showing that
these new methods provided better performance on large-scale data sets.

In this article, we will show that there exists a mathematically sound
explanation for the EPL approach. 
We will first introduce a likelihood function based on
ensemble learning, which marginalizes all possible building block
segmentations on the training data. Then, we will show that
the probability given by an EPL grammar is constant-factor
approximation of an ensemble method that integrates exponential number of models. 
Therefore, with an EPL grammar induction algorithm, we learn
a model with much more diversity from the training data. 
This may explain why EPL methods
provide state-of-the-art performance in many NLP pattern learning problems.

The rest of the article is organized as follows. 
We will first formalize EPL in Section \ref{sec:mono}.
In Section \ref{sec:bayes}, we introduce the ensemble method, and then
show the approximation theorem and its corollaries. 
We discuss a few important problems in Section \ref{sec:dis},
and conclude our work in Section \ref{sec:conc}. 

\section{Formalizing Exhaustive Pattern Learning} \label{sec:mono}

For the purpose of formalizing the core idea of EPL, we hereby
introduce a task called {\em monotonic translation}. 
Analysis on this task can be extended to other pattern
learning problems. Then, we will define segmentation on training data,
and introduce the EPL grammar, which will later be used in Section
\ref{sec:bayes}, theoretical justification of EPL.

\subsection{Monotonic Translation}

{\bf Monotonic translation} is defined as follows. The input
$\mathbf{x} \in {\mathcal{X}}$ is a string of words $x_1 x_2 ... x_i$ in 
the source language. The monotonic translation of $\mathbf{x}$ is
$\mathbf{y} \in {\mathcal{Y}}$, a string of words, $y_1 y_2 ... y_i$, of
the same length in the target language, where $y_j$ is the translation
of $x_j$, $1 \leq j \leq i$. 

In short, monotonic translation is a simplified version of machine
translation. There is no word reordering, insertion or deletion.
In this way, we ignore the impact of word level alignment, so as to
focus our effort on the study of building blocks. 
We leave the incorporation of
  alignments for future work. In fact, we can simply view alignments as
  constraints on building blocks.
  Monotonic translation is already general enough to model many NLP tasks such as
  labelling and chunking.

\subsection{Training Data Segmentation and MLE Grammars}

\begin{figure}[t]
\centering
   \scalebox{0.8}{\includegraphics{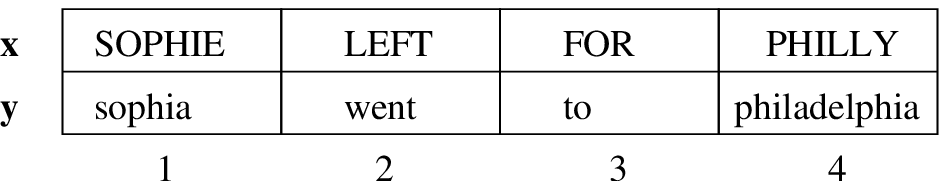}}
\caption{An example of training data for monotonic translation.\label{fig:train}}
\end{figure}

Without losing generality, we assume that the training data $D$ contains
a single pair of word strings, $\mathbf{x}_D$ and
${\mathbf{y}_D}$, which could be very long.
Let $\mathbf{x}_D$ = $x_1 x_2 ... x_n$, and $\mathbf{y}_D$ = $y_1 y_2
... y_n$. Source word $x_i$ is aligned to target word $y_i$.
Let the length of the word strings be $|D| = n$.
Figure \ref{fig:train} shows a simple example of training
data. Here $|D|$ = 4.

We assume that there exists a hidden segmentation on the training data,
which segments $\mathbf{x}_D$ and $\mathbf{y}_D$ into tokens. A {\bf token}
consists of a string of words, either on source or target, and it
contains at least one word. As for monotonic translation, the source
side and the target side share the same topology of segmentation.  
Tokens are the building blocks of the statistical model to be presented,
which means that the parameters for the model are defined on tokens
instead of words. 

A {\bf segmentation} $\mathbf{s}_D$ of $D$, or $\mathbf{s}$ for short,
is represented as a vector of $n-1$ Boolean values, $s_1 s_2 ... s_{n-1}$. 
$s_i = 0$ if and only if $x_i$ and $x_{i+1}$ belong to the same token.
$\mathbf{s}$ applies onto both the source and the target in the same
way, which means $x_i$ and $x_{i+1}$ belong to the same token if and
only if $y_i$ and $y_{i+1}$ belong to the same token.

\begin{figure}[t]
\begin{center}
   \scalebox{0.8}{\includegraphics{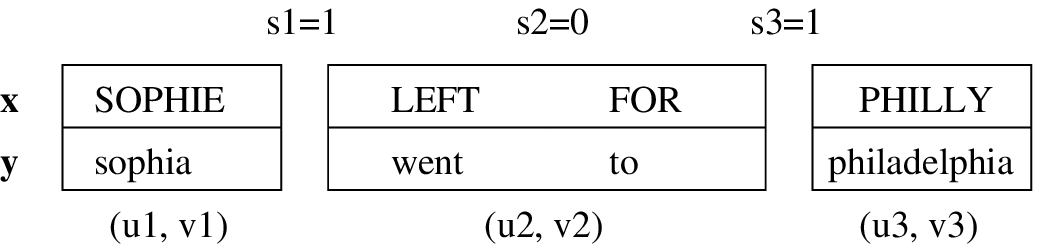}}
\caption{An example of segmentation on training data. \label{fig:seg}}
\end{center}
\end{figure}

If we segment $D$ with $\mathbf{s}$, we obtain a tokenized training set
${D_\mathbf{s}}$. ${D_\mathbf{s}}$ contains a pair of {\em token}
strings $\langle \mathbf{u_s}, \mathbf{v_s} \rangle$.
$\mathbf{u_s} = u_1 u_2 ... u_{|D_{\mathbf{s}}|}$, and 
$\mathbf{v_s} = v_1 v_2 ... v_{|D_{\mathbf{s}}|}$,
where $|D_{\mathbf{s}}|$ is the total number of tokens in $\mathbf{u_s}$
or $\mathbf{v_s}$.
Figure \ref{fig:seg} shows an example of segmentation on training
data. Here, $s_2=0$, so that we have a token pair that spans
two words, $(u_2,v_2) = (\mbox{LEFT FOR, went to})$.

Given training data $D$ and a segmentation $\mathbf{s}$ on $D$, there is a
unique joint probabilistic model obtained by the MLE on ${D_\mathbf{s}}$. 
Each parameter of this model contains a source token and target token.  
Since each token represents a string of words, we call this model a 
string-to-string {\bf grammar} $G_{{D\mathbf{s}}}$.
Specifically, for any pair of tokens $(u, v)$, we have  
\begin{eqnarray}
Pr(u, v | G_{{D\mathbf{s}}}) &=& \frac{\#_{\mathbf{s}}(u, v)}{|{D_\mathbf{s}}|}, \label{eqn:sgcnt}
\end{eqnarray}
where $\#_{\mathbf{s}}(u, v)$ is the number of times that this token pair 
appears in the segmented data ${D_\mathbf{s}}$.

As for the example segmentation $\mathbf{s}$ in Figure \ref{fig:seg}, its MLE grammar
is simply as follows. 
\begin{eqnarray}
Pr(\mbox{SOPHIE, sophia } | G_{{D\mathbf{s}}}) &=& 1/3 \nonumber \\ 
Pr(\mbox{LEFT FOR, went to }| G_{{D\mathbf{s}}}) &=& 1/3 \nonumber \\ 
Pr(\mbox{PHILLY, philadelphia }| G_{{D\mathbf{s}}}) &=& 1/3 \nonumber
\end{eqnarray}

However, for any given training data, its segmentation is unknown to
us. One way to cope with this problem is to consider all possible
segmentations. String distribution on the training data will lead us to
a good estimation of the hidden segmentation and tokens.
In Section \ref{sec:bayes}, we will introduce an ensemble
method to incorporate MLE grammars obtained from all
possible segmentations. Segmentations are generated with 
certain prior distribution.

\subsection{Exhaustive Pattern Learning for Monotonic Translation}

Now we present an EPL solution. We follow the widely-used heuristic method 
to generate a grammar by applying various
segmentations at the same time.
We build a heuristic grammar $G_{D,d}$ out of the training data $D$ by counting
all possible token pairs $(u, v)$
with at most $d$ words on each side, where
$d \ll |D|$ is a given parameter
\begin{eqnarray}
Pr(u, v | G_{D,d}) &=& \frac{\#(u, v)}{Z_d}, \nonumber
\end{eqnarray}
where $\#(u, v)$ is the number of times that the string pair encoded in
$(u, v)$ appears in $D$,\footnote{For the sake of convenience, in the
  rest of this article, we no longer distinguish a token and the string
  contained in this token unless necessary. We use symbols $u$ and $v$
  to represent both. The meaning is clear in context.}
and
$$Z_d = \sum_{(u', v')} \#(u', v') 
  = \sum_{i = 1...d} (|D|-i+1)
  = (1-\frac{d-1}{2|D|})d|D|.$$
Therefore,
\begin{eqnarray}
Pr(u, v | G_{D,d}) &=& \frac{\#(u, v)}{(1-\frac{d-1}{2|D|})d|D|}
\end{eqnarray}

For example, the heuristic grammar for the training data in Figure
\ref{fig:train} is as follows if we set $d=2$.
\begin{eqnarray}
Pr(\mbox{SOPHIE, sophia } | G_{D,2}) &=& 1/7\nonumber \\
Pr(\mbox{LEFT, went } | G_{D,2}) &=& 1/7\nonumber \\
Pr(\mbox{FOR, to } | G_{D,2}) &=& 1/7\nonumber \\
Pr(\mbox{PHILLY, philadelphia } | G_{D,2}) &=& 1/7\nonumber \\
Pr(\mbox{SOPHIE LEFT, sophia went } | G_{D,2}) &=& 1/7\nonumber \\
Pr(\mbox{LEFT FOR, went to } | G_{D,2}) &=& 1/7\nonumber \\
Pr(\mbox{FOR PHILLY, to philadelphia } | G_{D,2}) &=& 1/7\nonumber
\end{eqnarray}
A desirable translation rule `LEFT FOR $\Rightarrow$ went to' is in this
heuristic grammar, although its weight is diluted by noise. The hope is
that, good translation rules will appear more often in the training
data, so that they can be distinguished from noisy rules.

In the decoding phase, we use grammar $G_{D,d}$ as if it is a regular
MLE grammar.
Let $\mathbf{x}$ = $x_1x_2...x_i$ be an input source string. For any segmentation 
$\mathbf{a}$ on the {\em test} sentence $\mathbf{x}$, 
let $\mathbf{u_a} = u_1u_2...u_k$ be the resultant string
of source tokens. The length of the string is $|\mathbf{x}|=i$, and the
length of the token string is $|\mathbf{u_a}| = k$.
The translation that we are looking for is given
by the target token vector $\hat{\mathbf{v}}$, such that 
\begin{eqnarray}
\langle \hat{\mathbf{v}}, \hat{\mathbf{a}} \rangle 
&=& \argmax_{\langle \mathbf{v}, \mathbf{a} \rangle}
Pr(\mathbf{u_a},\mathbf{v} | G_{D,d}), \mbox{ where} \nonumber \\
Pr(\mathbf{u_a},\mathbf{v} | G_{D,d}) &=& \prod_{j=1...|\mathbf{u_a}|} Pr(u_j,v_j | G_{D,d}) \\
&=& \prod_{j=1...|\mathbf{u_a}|} \frac{m_j}{(1-\frac{d-1}{2|D|})d|D|} \nonumber
\end{eqnarray}
where $m_j = \#(u_j, v_j)$.
As in previous work of structure-based MT, 
we do not calculate the marginal probability that sums up all possible
target tokens generating the same word string, due to the concern
of computational complexity.

Obviously, with $G_{D,d}$, we can take advantage of larger context
of up to $d$ words. However, a common criticism against
the EPL approach is that a grammar like $G_{D,d}$ is not mathematically
sound. The probabilities are simply heuristics, and there is no clear statistical explanation.
In the next section, we will show that $G_{D,d}$ is mathematically sound.

\section{Theoretical Justification of Exhaustive Pattern Learning} \label{sec:bayes}

In this section, we will first introduce an ensemble model and a prior distribution of segmentation.
Then we will show the theorem of approximation, and present corollaries on conditional probabilities and tree structures.

\subsection{An Ensemble Model}

Let $D$ be the training data of $|D|$ words. Let $\mathbf{s}$ be an
arbitrary token segmentation on $D$, where $\mathbf{s}$ is unknown to
us. Given $D$ and $\mathbf{s}$, we can obtain a model/grammar
$G_{D\mathbf{s}}$ with maximum likelihood estimation. Thus, we can
calculate joint probability of $(u_j, v_j)$ given grammar
$G_{D\mathbf{s}}$, $Pr(u_j,v_j | G_{D\mathbf{s}})$.

There are potentially exponential number of distinct segmentations for
$D$. Here, we use an ensemble method to sum over all possible
segmentations. This method would provide desirable coverage and diversity of
translation rules to be learned from the training data. For each segmentation
$\mathbf{s}$, we have a fixed prior probability $Pr(\mathbf{s})$ which
we will shown in Section \ref{sec:prior}. Thus, we define the ensemble
probability $L(u_j, v_j)$ as follows.

\begin{eqnarray}
L(u_j, v_j) &=& \sum_{\mathbf{s}} Pr(u_j, v_j | G_{{D\mathbf{s}}}) Pr(\mathbf{s}). \label{eqn:bayes}
\end{eqnarray}

Prior segmentation probabilities $Pr(\mathbf{s})$ serve as model
probabilities in (\ref{eqn:bayes}). Having the model probabilities fixed
in this way could avoid over-fitting of the training data \cite{DGZ06}.

In decoding, we search for the best hypothesis $\hat{\mathbf{v}}$ given training data $D$ and input $\mathbf{x}$ as follows.
\begin{eqnarray}
\langle \hat{\mathbf{v}}, \hat{\mathbf{a}} \rangle &=& \argmax_{\langle
  \mathbf{v}, \mathbf{a} \rangle} L(\mathbf{u_a},\mathbf{v}), \mbox{ where} \nonumber \\
L(\mathbf{u_a},\mathbf{v}) &=& \prod_{j=1...|\mathbf{u_a}|} L(u_j, v_j) \nonumber
\end{eqnarray}


What is interesting is that there turns out to be a prior
distribution for $\mathbf{s}$, such that, under certain conditions, the
limit of $L(\mathbf{u_a},\mathbf{v}) / Pr(\mathbf{u_a},\mathbf{v} | G_{D,d})$
as $|D|\to\infty$ is a value that depends only on $|\mathbf{x}|$ and a
parameter of the prior distribution $Pr(\mathbf{s})$, to be shown in Theorem \ref{thm:bayes}. 
$|\mathbf{x}|$ is a constant for all hypotheses for the same input. Therefore,
$Pr(\mathbf{u_a},\mathbf{v} | G_{D,d})$ is constant-factor approximation of
$L(\mathbf{u_a},\mathbf{v})$. Using $G_{D,d}$ is, to some extent, equivalent
to using all possible MLE grammars at the same time via an ensemble method.

\subsection{Prior Distribution of Segmentation} \label{sec:prior}

Now we define a probabilistic model to generate segmentation. 
$\mathbf{s} = \langle s_1, s_2, ... , s_{|D|-1}\rangle$ is
a vector of $|D|-1$ independent Bernoulli variables.
$s_i$ represents whether $x_i$ and $x_{i+1}$ belong to separated tokens. 
1 means yes and 0 means no. All
the individual separating variables have the same distribution,
$P_q(s_i=0)=q$ and $P_q(s_i=1)=1-q$, for a given real value $q$, $0
\leq q \leq 1$. Since $L(\mathbf{u_a},\mathbf{v})$ depends on $q$ now, we
rewrite it as $L_q(\mathbf{u_a},\mathbf{v})$.

Based on the definition, Lemma \ref{lem:one} immediately follows, which
will be used later.
\begin{lemma} \label{lem:one}
For each string pair $(u, v)$, the probability that an appearance of
$(u, v)$ in $D$ is exactly tokenized as $u$ and $v$ by $\mathbf{s}$ 
is $q^{|u|-1}(1-q)^2$.
\end{lemma}


\subsection{Theorem of Approximation}

Let $\mathbf{x}$ = $x_1x_2...x_i$ be an input source string. Let
$\mathbf{a}$ be a segmentation on $\mathbf{x}$, and the resultant token
string be $\mathbf{u_a} = u_1u_2...u_k$. Let $\mathbf{v} = v_1v_2...v_k$ be a
hypothesis translation of $\mathbf{u_a}$. Let $m_j = \#(u_j, v_j)$, the number of times that
string pair $(u_j, v_j)$ appears in the training data $D$, $1\leq j \leq k$.
Let $m_{j,\mathbf{s}} = \#_{\mathbf{s}}(u_j, v_j)$, the number of times
that this token pair appears in the segmented data ${D_\mathbf{s}}$.
In order to prove Theorem \ref{thm:bayesT}, we assume that the following two
assumptions are true for any pair of tokens $(u_j,v_j)$.
\begin{assumption} \label{asm:one}
Any two of the $m_j$ appearances in $D$ are neither overlapping nor consecutive.
\end{assumption}

This assumption is necessary for the calculation of $\mbox{E}[m_{j,\mathbf{s}}]$, $1 \leq j \leq k$.
Based on Lemma \ref{lem:one},
the number of times that $(u_j, v_j)$ is exactly tokenized as in this way
with segmentation $\mathbf{s}$ is in a binomial distribution $B(m_j,
q^{|u_j|-1}(1-q)^2)$, so that
\begin{eqnarray} \label{eqn:emj}
\mbox{E}[m_{j,\mathbf{s}}] &=& m_jq^{|u_j|-1}(1-q)^2, \nonumber
\end{eqnarray}
where $|u_j|$ is the number of words in $u_j$. In addition, since there
is no overlap, these appearances cover a total of $|u_j|m_j$ source words. 

\begin{assumption} \label{asm:two}
Let $\eta_j = \frac{(|u_j|+1)m_j}{|D|}$. We have
$\lim_{|D|\to\infty} \eta_j = 0$.
\end{assumption}

In fact, as we will see it in Section \ref{sec:disasm}, we do not have to rely on Assumption
\ref{asm:two} to bound the ratio of $Pr(\mathbf{u_a},\mathbf{v}
| G_{D,d})$ and $L_q(\mathbf{u_a},\mathbf{v})$. We know that
$\eta_j$ is a very small positive number,
and we can build the upper and lower bounds of the ratio based on
$\eta_j$. However, with this assumption, it will be
much easier to see the big picture, so we assume that it is true in the
rest of this section. 

\begin{theorem} \label{thm:bayesT} 
Suppose Assumptions \ref{asm:one} and \ref{asm:two} hold for a given
pair of tokens $(u_j, v_j)$, then we have
\begin{eqnarray}
\lim_{|D|\to\infty}
\frac{L_{q}(u_j,v_j)}{Pr(u_j,v_j| G_{D,d})} &=& q^{|u_j|}, \nonumber
\end{eqnarray}
where $q = d/(d+1)$.
\end{theorem}

Later in the section, we will show Theorem \ref{thm:bayesT} with Lemmas \ref{lem:rit} and \ref{lem:lft}.
Theorem \ref{thm:bayes} immediately follows Theorem \ref{thm:bayesT}.

\begin{theorem} \label{thm:bayes}
Suppose Assumptions \ref{asm:one} and \ref{asm:two} hold for any $j$, then we have
\begin{eqnarray}
\lim_{|D|\to\infty}
\frac{L_{q}(\mathbf{u_a},\mathbf{v})}{Pr(\mathbf{u_a},\mathbf{v}
  | G_{D,d})} &=& q^{|\mathbf{x}|}, \nonumber
\end{eqnarray}
where $q = d/(d+1)$.
\end{theorem}
Here, $|\mathbf{x}|$ is a constant for hypotheses of the same input.
An interesting observation is that the prior segmentation model to
fit into this theorem tends to generate longer tokens, if we have a
larger value for $d$.

We will show Theorem \ref{thm:bayesT} by bounding it from above and
below via Lemmas \ref{lem:rit} and \ref{lem:lft} respectively. Now, we
introduce the notations to be use the proofs of Lemmas \ref{lem:rit} and
\ref{lem:lft} in Appendixes A and B respectively.

First, we combine (\ref{eqn:sgcnt}) and (\ref{eqn:bayes}), and obtain
\begin{eqnarray}
L_q(u_j, v_j) 
  &=& E_{\mathbf{s}} (\frac{m_{j,\mathbf{s}}}{|{D_\mathbf{s}}|}).
    \label{eqn:expt1}
\end{eqnarray}
With Assumption \ref{asm:one}, we know the value of $\mbox{E}[m_{j,\mathbf{s}}]$.
However, $|{D_\mathbf{s}}|$ depends on
$m_{j,\mathbf{s}}$, and this prevents us from computing the expected value on each
individual item.

\begin{figure}[t]
\centering
   \scalebox{0.8}{\includegraphics{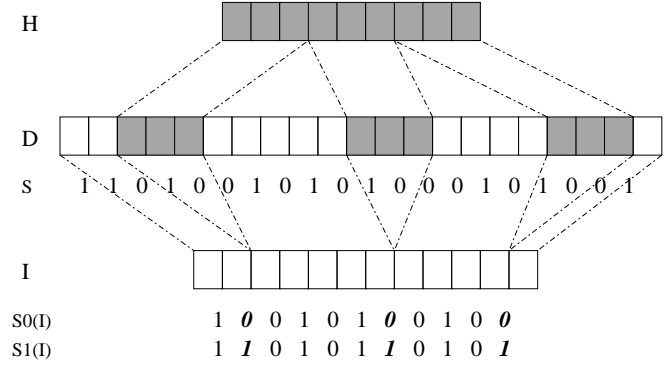}}
\caption{An example of training data splitting.\label{fig:hi}}
\end{figure}

We solve it by bounding $|{D_\mathbf{s}}|$ with
values independent of $m_{j,\mathbf{s}}$, or the separating variables
related to the $m_j$ appearances in $D$.
We divide $D$ into two parts, $H$ and $I$, based on the $m_j$ appearances
of $(u_j,v_j)$ pairs. 
$H$ is the part that contains and only contains
all $m_j$ appearances, and $I$ is the rest of $D$, 
so that the internal separating
variables of $I$ are independent of $m_{j,\mathbf{s}}$.
An example is shown in Figure \ref{fig:hi}.
Black boxes represent the $m_j$ appearances.

We concatenate fragments in $I$ and
keep the $I$-internal separating variables as in $\mathbf{s}$. 
There are two variants of the segmentation for $I$, depending on how we
define the separating variables between the fragments. So we have the following two
segmented sub-sets.
\begin{itemize}
\item $I_{\mathbf{s}_0(I)}$: inter-fragment separating variable = 0. 
\item $I_{\mathbf{s}_1(I)}$: inter-fragment separating variable = 1. 
\end{itemize}
Here, $\mathbf{s}_0(I)$ and $\mathbf{s}_1(I)$ represent the two
segmentation vectors on $I$ respectively,
each of which has $|I|- m_j - 1$ changeable separating variables,
where $|I|$ is the number words contained in $I$.
The number of changeable variables that set to $1$ follows a
binomial distribution $B(|I|- m_j - 1, 1-q)$.
In Figure \ref{fig:hi}, fixed inter-fragment separating variables are
represented in the bold italic font.

If there are $s$ changeable variables set to $1$ in $I_{\mathbf{s}_0(I)}$, 
the number of tokens in $I_{\mathbf{s}_0(I)}$ is $|I_{\mathbf{s}_0(I)}|=s+1$. 
Similarly, if there are $s$ changeable variables set to $1$ in $I_{\mathbf{s}_1(I)}$,
the number of tokens in $I_{\mathbf{s}_1(I)}$ is $|I_{\mathbf{s}_1(I)}| = s+m_j+1$. 

In addition, it is easy to verify that
\begin{eqnarray} 
|I_{\mathbf{s}_0(I)}| &\leq& |D_\mathbf{s}| \label{eqn:i0} \\
|D_\mathbf{s}| &\leq& |I_{\mathbf{s}_1(I)}| + |u_j| m_j  \label{eqn:i1}
\end{eqnarray}


Combining (\ref{eqn:expt1}), (\ref{eqn:i0}) and the two assumptions,
we have the upper bound in Lemma \ref{lem:rit}.
\begin{lemma} \label{lem:rit}
If Assumptions \ref{asm:one} and \ref{asm:two} hold,
$$\lim_{|D|\to\infty} \frac{L_q(u_j,v_j)}{Pr(u_j,v_j | G_{D,d})} \leq q^{|u_j|},$$ where $q = d/(d+1)$.
\end{lemma}

Similarly, combining (\ref{eqn:expt1}), (\ref{eqn:i1}) and the two assumptions,
we obtain the lower bound in Lemma \ref{lem:lft}.
\begin{lemma} \label{lem:lft}
If Assumptions \ref{asm:one} and \ref{asm:two} hold,
$$\lim_{|D|\to\infty} \frac{L_q(u_j,v_j)}{Pr(u_j,v_j | G_{D,d})} \geq q^{|u_j|},$$ where $q = d/(d+1)$.
\end{lemma}

The proofs of Lemmas \ref{lem:rit} and \ref{lem:lft} are given in the
Appendixes A and B respectively. 
The proof of Lemma \ref{lem:rit} also depends on Lemma \ref{lem:inv}.
Lemma \ref{lem:inv} and its proof are given in Appendix C.

Therefore, Theorem \ref{thm:bayesT} holds. 


\subsection{Corollaries on Conditional Probabilities}

Theorem \ref{thm:bayesT} is for joint distribution of token pairs. In
previous work of using EPL, conditional probabilities were ofter used,
for example, like $P(u|v)$ and $P(v|u)$. Starting from Theorem
\ref{thm:bayesT}, we can easily obtain the following corollaries for
conditional probabilities.

\begin{corollary} \label{col:uv}
Suppose Assumptions \ref{asm:one} and \ref{asm:two} hold for a given
pair of tokens $(u_j, v_j)$, then we have
\begin{eqnarray}
\lim_{|D|\to\infty} Pr(u_j | v_j, G_{D,d}) / \frac{L_q(u_j,v_j)}{\sum_u L_q(u, v_j)} &=& 1, \nonumber
\end{eqnarray}
where $q = d/(d+1)$.
\end{corollary}

{\bf Proof}
According to the definition, $Pr(u, v_j | G_{D,d}) = 0$ and $L_q(u, v_j)
= 0$, if $|u| \neq |v_j|$.
Therefore, we only need to consider all pairs of $(u, v_j)$, such that $|u| = |v_j| =
|u_j|$. The number of distinct $u$ is a finite number, since source
vocabulary is a finite set. Therefore, according to Theorem
\ref{thm:bayesT}, for any small positive number $\epsilon$, there exists
a positive number $n$, such that, if $|D| > n$, we have
\begin{eqnarray} \label{eqn:conb}
(1-\epsilon)\frac{L_q(u, v_j)}{q^{|u_j|}} \leq Pr(u, v_j | G_{D,d}) \leq
(1+\epsilon)\frac{L_q(u, v_j)}{q^{|u_j|}}
\end{eqnarray}
Therefore, we have
\begin{eqnarray}
Pr(u_j | v_j, G_{D,d}) 
&=& \frac{Pr(u_j, v_j | G_{D,d})}{\sum_{u:|u|=|u_j|} Pr(u, v_j | G_{D,d})} \nonumber\\
&\leq& \frac{(1+\epsilon)L_q(u,v_j)/q^{|u_j|}}{\sum_{u:|u|=|u_j|} (1-\epsilon) L_q(u,v_j)/q^{|u_j|}} \mbox{ \{Eqn. (\ref{eqn:conb})\}}\nonumber \\
&=& \frac{1+\epsilon}{1-\epsilon} \frac{L_q(u_j, v_j)}{\sum_u L_q(u,v_j)} \nonumber
\end{eqnarray}
Thus,
\begin{eqnarray}
Pr(u_j | v_j, G_{D,d}) / \frac{L_q(u_j,v_j)}{\sum_u L_q(u, v_j)} &\leq& \frac{1+\epsilon}{1-\epsilon} \nonumber
\end{eqnarray}
Similarly, we have
\begin{eqnarray}
Pr(u_j | v_j, G_{D,d}) / \frac{L_q(u_j,v_j)}{\sum_u L_q(u, v_j)} &\geq& \frac{1-\epsilon}{1+\epsilon} \nonumber
\end{eqnarray}
Therefore, $$\lim_{|D|\to\infty} Pr(u_j | v_j, G_{D,d}) / \frac{L_q(u_j,v_j)}{\sum_u L_q(u, v_j)} = 1$$

\begin{corollary} \label{col:vu}
Suppose Assumptions \ref{asm:one} and \ref{asm:two} hold for a given
pair of tokens $(u_j, v_j)$, then we have
\begin{eqnarray}
\lim_{|D|\to\infty} Pr(v_j | u_j, G_{D,d}) / \frac{L_q(u_j,v_j)}{\sum_v
  L_q(u_j, v)} &=& 1, \nonumber
\end{eqnarray}
where $q = d/(d+1)$.
\end{corollary}


The proof of Corollary \ref{col:vu} is similar to that of Corollary \ref{col:uv}.
Therefore, conditional probabilities in EPL are reasonable
approximation of the conditional ensemble probability functions.

The proofs for the conditional probabilities depend on a special
property of monotonic translation; the length of $u_j$ is the same as
the length of $v_j$. However, this is not true in real application of
phrase-based translation. The source and the target sides may have
different segmentations. We leave the modeling of real phrase-based
translation for future work.

\subsection{Extension to Tree Structures} \label{sec:es2t}

Now we try to extend Theorem \ref{thm:bayesT} to the string-to-tree grammar. 
First, we define a prior distribution on tree segmentation. 
We assign a Bernoulli variable to each tree node,
representing the probabilities that we separate the tree at this
node, i.e, with probabilities of $1-q$, we choose to separate
each node. 

Let $(u_j, v_j)$ be a string--tree pair, where $u_j$ is a source string and
$v_j$ is a target tree.
Let $t_j$ be the number of words
in $u_j$, and let $n_j$ be the number of non-terminals in $u_j$, where
$t_j + n_j \leq d$, and $\sum t_j$ is the length of the input sentence, $|\mathbf{x}|$. 
Thus, the probability that an appearance of $(u_j, v_j)$ in $D$ is
exactly tokenized as in this way is $q^{t_j-1}(1-q)^{n_j+1}$. 

With similar methods used in the proofs for string structures, we can
show that, if Assumptions
\ref{asm:one} and \ref{asm:two} hold, 
\begin{eqnarray}
\lim_{|D|\to\infty} \frac{L_q(u_j,v_j)}{Pr(u_j,v_j | G_{D,d})} 
&=& c \, q^{t_j-1}(1-q)^{n_j}, \nonumber
\end{eqnarray}
where $c = \sum_{i:\,t_i+n_i \leq d}\#(u_i,v_i)/|D|$ is a constant, and $q$ is a free
parameter. We skip the proof here to avoid duplication of similar procedure. 
We define
\begin{eqnarray}
P_{q,d}(u_j,v_j) &=& c \, q^{t_j-1}(1-q)^{n_j} Pr(u_j,v_j | G_{D,d}).\nonumber
\end{eqnarray}
Thus, $P_{q,d}(u_j,v_j)$ approximates $L_q(u_j, v_j)$,
where
\begin{eqnarray}
\lim_{|D|\to\infty} \frac{L_q(u_j,v_j)}{P_{q,d}(u_j,v_j)} = 1. \nonumber
\end{eqnarray}
This result shows a theoretically better way of using heuristic grammar
in string-to-tree models.

\section{Discussion} \label{sec:dis}

In this section, we will focus on three facts that need more explanation.

\subsection{On the Use of Assumption \ref{asm:two}}  \label{sec:disasm}

In the proofs of Lemmas \ref{lem:rit} and \ref{lem:lft},
Assumption \ref{asm:two} is only used in the very last steps.
Therefore, we could build the upper and lowers bounds of the ratio
without Assumption \ref{asm:two} by connecting Inequalities
(\ref{eqn:ubt}) and (\ref{eqn:bbt}) in Appendixes A and B respectively.

\subsection{On the Ensemble Probability}  \label{sec:disens}

The ensemble probability in (\ref{eqn:bayes}) can be viewed as
simplification of a Bayesian model in (\ref{eqn:bayesG}). 
\begin{eqnarray}
L(u_j, v_j) &=& Pr(u_j, v_j | D) \nonumber \\
 &=& \sum_{G\in\mathcal{G}(D)} Pr(u_j, v_j | G) Pr(G|D) \label{eqn:bayesG}
\end{eqnarray}
In (\ref{eqn:bayesG}), we marginalize all possible token-based grammars $G$
from $D$, $\mathcal{G}(D)$. Furthermore,
\begin{eqnarray}
Pr(G|D) &=&\sum_{\mathbf{s}} Pr(G|D,\mathbf{s}) Pr(\mathbf{s}|D)\nonumber
\end{eqnarray}
Then, we approximate the posterior probability of $G$ given $D$ and $\mathbf{s}$ with point
estimation. Thus, $Pr(G|D,\mathbf{s}) = 1$ if and only if $G$ is the MLE grammar
of $D\mathbf{s}$, which means
all the distribution mass is assigned to $G_{D\mathbf{s}}$, the MLE grammar for $D\mathbf{s}$.
We also assume that $\mathbf{s}$ is independent of $D$. Thus, 
\begin{eqnarray}
Pr(G|D) &=&\sum_{\mathbf{s}}\mathbf{1}(G=G_{D\mathbf{s}}) Pr(\mathbf{s}), \label{eqn:one}
\end{eqnarray}
where $Pr(\mathbf{s})$ is a prior distribution of segmentation for any string of $|D|$ words.
With (\ref{eqn:one}), we can rewrite (\ref{eqn:bayesG}) as follows.
\begin{eqnarray}
L(u_j, v_j)
&=& \sum_{G\in\mathcal{G}(D)} Pr(u_j, v_j | G) 
                \sum_{\mathbf{s}}\mathbf{1}(G=G_{D\mathbf{s}}) Pr(\mathbf{s})\nonumber \\
&=& \sum_{\mathbf{s}} \sum_{G\in\mathcal{G}(D)} 
    Pr(u_j, v_j | G) \mathbf{1}(G=G_{D\mathbf{s}}) Pr(\mathbf{s}) \nonumber \\
&=& \sum_{\mathbf{s}} Pr(u_j, v_j | G_{{D\mathbf{s}}}) Pr(\mathbf{s}) \label{eqn:bayes2}
\end{eqnarray}
Equation (\ref{eqn:bayes2}) is exactly the ensemble probability in
Equation (\ref{eqn:bayes}).

\subsection{On the DOP Model}  \label{sec:disdop}

The EPL method investigated in this article may date back to Data
Oriented Parsing (DOP) by \cite{Bod92}. What is special with DOP is that
the DOP model uses overlapping treelets of various sizes in an
exhaustive way as building blocks of a statistical tree grammar. 

In our framework, for each pair $(u_j, v_j)$, we can use $u_j$ to represent
the input text, and $v_j$ to represent its tree structure. Thus, it
would be similar to the string-to-tree model in Section
\ref{sec:es2t}. Joint probability of $(u_j, v_j)$ stands for unigram probability
$Pr(\mbox{treelet})$. 

However, the original DOP estimator (DOP1) is quite different from our
monotonic translation model. The conditional probability
in DOP1 is defined as $Pr(\mbox{treelet} | \mbox{subroot-label})$, so that there is no
obvious way to model DOP1 with monotonic translation.
Therefore, theoretical justification of DOP1 is still an open problem.

\section{Conclusion} \label{sec:conc}
In this article, we first formalized exhaustive pattern learning (EPL),
which is widely used in grammar induction in NLP.  
We showed that using an EPL heuristic grammar is equivalent to
using an ensemble method to cope with the uncertainty of building blocks
of statistical models. 

Better understanding of EPL may lead to improved pattern learning
algorithms in future. This work will affect the research in various
fields of natural language processing, including machine translation,
parsing, sequence classification etc.
EPL can also been applied to other research fields outside NLP.

\section*{Acknowledgments}
This work was inspired by enlightning discussion with Scott Miller, 
Rich Schwartz and Spyros Matsoukas when the author was at BBN Technologies.
Reviewers of ACL 2010, CoNLL 2010, EMNLP 2010, ICML 2011, CLJ and JMLR helped 
to sharpen the focus of this work. However, all the mistakes belong to me.

\section*{Appendix A. Proof for Lemma \ref{lem:rit}}

\begin{eqnarray}
L_q(u_j, v_j) 
&=& \mbox{E}_{\mathbf{s}} [\frac{m_{j,\mathbf{s}}}{|{D_\mathbf{s}}|}] \mbox{ \{Eqn. (\ref{eqn:expt1}) \}} \nonumber \\
&\leq& \mbox{E}_{\mathbf{s}} [\frac{m_{j,\mathbf{s}}}{|I_{\mathbf{s}_0(I)}|}] \mbox{ \{Eqn. (\ref{eqn:i0}) \}} \nonumber \\
&=& \mbox{E}[m_{j,\mathbf{s}}] \mbox{E}[\frac{1}{|I_{\mathbf{s}_0(I)}|}] \nonumber \\ 
  &&\mbox{ \{Independence of }m_{j,\mathbf{s}}\mbox{ and }I_{\mathbf{s}0(I)}\} \nonumber \\
&=& \mbox{E}[m_{j,\mathbf{s}}] \mbox{E}_{\mathbf{s}_0(I)}[\frac{1}{|I_{\mathbf{s}0(I)}|}] \nonumber \\
&=& \frac{\mbox{E}[m_{j,\mathbf{s}}]}{(1-q)(|I|-m_j)}(1-q^{|I|-m_j}) \mbox{ \{Lemma \ref{lem:inv}\}}  \nonumber \\
&\leq& \frac{\mbox{E}[m_{j,\mathbf{s}}]}{(1-q)(|I|-m_j)} \nonumber \\
&=& \frac{m_jq^{|u_j|-1}(1-q)^2}{(1-q)(|I|-m_j)} \mbox{ \{Binomial Dist., Assumption \ref{asm:one}\}} \nonumber \\ 
&=& \frac{m_jq^{|u_j|-1}(1-q)^2}{(1-q)(|D|-|u_j|m_j - m_j)}  \nonumber \\
&=& \frac{m_jq^{|u_j|-1}(1-q)^2}{(1-q)(1-\eta_j)|D|} \nonumber \\
&=& \frac{m_j}{(1-\frac{d-1}{2|D|})d|D|} \frac{q^{|u_j|-1}(1-q)^2(1-\frac{d-1}{2|D|})d}{(1-q)(1-\eta_j)} \nonumber \\
&=& Pr(u_j,v_j | G_{D,d}) \frac{q^{|u_j|-1}(1-q)^2(1-\frac{d-1}{2|D|})d}{(1-q)(1-\eta_j)} \nonumber \\
&=& Pr(u_j,v_j | G_{D,d}) q^{|u_j|} \frac{(1-q)(1-\frac{d-1}{2|D|})d}{(1-\eta_j)q} \label{eqn:ubt}
\end{eqnarray}
\begin{eqnarray}
&&\lim_{|D|\to\infty} \frac{L_q(u_j,v_j)}{Pr(u_j,v_j | G_{D,d})} \nonumber \\
&\leq& q^{|u_j|} \frac{(1-q)d}{q} \mbox{ \{Assumption \ref{asm:two}\}} \nonumber \\ 
&=& q^{|u_j|}. \nonumber
\end{eqnarray}

\newpage

\section*{Appendix B. Proof for Lemma \ref{lem:lft}}

\begin{eqnarray}
L_q(u_j,v_j)
&=& \mbox{E}_{\mathbf{s}} [\frac{m_{j,\mathbf{s}}}{|{D_\mathbf{s}}|}] \mbox{ \{Eqn. (\ref{eqn:expt1}) \}} \nonumber \\
&\geq& \mbox{E}_{\mathbf{s}} [\frac{m_{j,\mathbf{s}}}{|I_{\mathbf{s}_1(I)}| + |u_j|m_j}] \mbox{ \{Eqn. (\ref{eqn:i1})\}} \nonumber \\
&=& \mbox{E}[m_{j,\mathbf{s}}]\mbox{E}[\frac{1}{|I_{\mathbf{s}_1(I)}| + |u_j|m_j}] \nonumber \\
  &&\mbox{ \{Independence of }m_{j,\mathbf{s}}\mbox{ and } I_{\mathbf{s}1(I)}\} \nonumber \\
&\geq& \frac{\mbox{E}[m_{j,\mathbf{s}}]}{\mbox{E}[|I_{\mathbf{s}_1(I)}| + |u_j|m_j]} \mbox{\{Jensen's inequality\}}  \nonumber \\
&=& \frac{\mbox{E}[m_{j,\mathbf{s}}]}{\mbox{E}_{\mathbf{s}_1(I)}[|I_{\mathbf{s}_1(I)}|] + |u_j|m_j} \nonumber \\ 
&=& \frac{m_jq^{|u_j|-1}(1-q)^2}{(1-q)(|I|-1-m_j)+m_j + 1 + |u_j|m_j} \nonumber \\
  &&\mbox{ \{Binomial Dist., Assumption \ref{asm:one}\}} \nonumber \\ 
&=& \frac{m_jq^{|u_j|-1}(1-q)^2}{(1-q)(|D|-|u_j|m_j)+q(1+m_j) + |u_j|m_j}\nonumber \\
&=& \frac{m_jq^{|u_j|-1}(1-q)^2}{(1-q)|D|+q(|u_j|m_j + m_j) + q}  \nonumber \\
&=& \frac{m_jq^{|u_j|-1}(1-q)^2}{(1-q+q\eta_j+\frac{q}{|D|})|D|}\nonumber \\
&=& \frac{m_j}{(1-\frac{d-1}{2|D|})d|D|} \frac{q^{|u_j|-1}(1-q)^2(1-\frac{d-1}{2|D|})d}{1-q+q\eta_j+\frac{q}{|D|}} \nonumber \\
&=& Pr(u_j,v_j | G_{D,d}) \frac{q^{|u_j|-1}(1-q)^2(1-\frac{d-1}{2|D|})d}{1-q+q\eta_j+\frac{q}{|D|}}  \nonumber \\
&=& Pr(u_j,v_j | G_{D,d}) q^{|u_j|} \frac{(1-q)^2(1-\frac{d-1}{2|D|})d}{(1-q+q\eta_j+\frac{q}{|D|})q} \label{eqn:bbt}
\end{eqnarray}
\begin{eqnarray}
&&\lim_{|D|\to\infty} \frac{L_q(u_j,v_j)}{Pr(u_j,v_j | G_{D,d})} \nonumber \\
&\geq& q^{|u_j|} \frac{(1-q)d}{q} \mbox{ \{Assumption \ref{asm:two}\}} \nonumber \\ 
&=& q^{|u_j|}. \nonumber
\end{eqnarray}

\newpage

\section*{Appendix C. Lemma \ref{lem:inv} and its Proof}

\begin{lemma} \label{lem:inv}
Let $X$ be a random variable of Binomial distribution $B(n, 1-q)$, then
$$\mbox{E}[\frac{1}{X+1}] = \frac{1-q^{n+1}}{(1-q)(n+1)}$$
\end{lemma}
\begin{eqnarray}
\mbox{E}[\frac{1}{X+1}]
&=& \sum_{k=0}^{n} \frac{1}{k+1} \frac{n!}{k!(n-k)!} q^{n-k} (1-q)^k \nonumber \\
&=& \frac{1}{(1-q)(n+1)} \nonumber \\
  &&\sum_{k=0}^{n} \frac{(n+1)!}{(k+1)!(n-k)!} q^{n-k} (1-q)^{k+1} \nonumber \\
&=& \frac{1}{(1-q)(n+1)} ((q+1-q)^{n+1}-q^{n+1}) \nonumber \\
&=& \frac{1-q^{n+1}}{(1-q)(n+1)} \nonumber
\end{eqnarray}

\bibliography{ainlp}
\bibliographystyle{fullname}

\end{document}